# Rerendering Semantic Ontologies:
# Automatic Extensions to UMLS through Corpus Analytics

**James Pustejovsky, Anna Rumshisky, José Castaño**

Department of Computer Science, Brandeis University
Waltham, MA 02454
{jamesp, arum, jcastano}@cs.brandeis.edu

**Abstract**
In this paper, we discuss the utility and deficiencies of existing ontology resources for a number of language processing applications. We describe a technique for increasing the semantic type coverage of a specific ontology, the National Library of Medicine's UMLS, with the use of robust finite state methods used in conjunction with large-scale corpus analytics of the domain corpus. We call this technique "semantic rendering" of the ontology. This research has been done in the context of Medstract, a joint Brandeis-Tufts effort aimed at developing tools for analyzing biomedical language (i.e., Medline), as well as creating targeted databases of bio-entities, biological relations, and pathway data for biological researchers (Pustejovsky et al., 2002). Motivating the current research is the need to have robust and reliable semantic typing of syntactic elements in the Medline corpus, in order to improve the overall performance of the information extraction applications mentioned above.

## 1. Introduction

Data mining and information extraction rely on a number of natural language tasks that require semantic typing; that is, the ability of an application to accurately determine the conceptual categories of syntactic constituents. Accurate semantic typing serves tasks such as relation extraction by improving anaphora resolution and entity identification. Domain-specific semantic typing also benefits statistical categorization and disambiguation techniques that require generalizations across semantic classes to make up for the sparsity of data. This applies, for example, to the problem of prepositional attachment, as well as identification of semantic relations between constituents within nominal compounds (see, for example, related discussion in Rosario & Hearst (2002)). Semantic typing has other direct applications, such as query reformulation, the filtering of results according to semantic type restrictions, and so on.

The set of categories used in semantic typing must be adequate enough to serve such tasks. In the biomedical domain, there are a number of efforts to develop specialized taxonomies and knowledge bases (UMLS, Gene Ontology, SWISS-PROT, OMIM, DIP). In this paper, we describe a method for adapting existing ontology resources for the natural language processing tasks and illustrate this technique on the National Library of Medicine's UMLS.

The UMLS, like many industry-standard taxonomies, contains a large number of word-concept pairings (over 1.5M typed terms), making it potentially attractive as a resource for semantic tagging information. However, these types are inadequate for NL tasks for two major reasons. First, the overall type structure is very shallow. For example, for the semantic tag "Amino Acid, Peptide, or Protein" (henceforth AAPP), there are 180,998 entries, for which there are dozens of functional subtypes that are routinely distinguished by biologists, but not in the UMLS.

One specific example of the type system deficiencies illustrates this point very clearly: the extraction of relations and their arguments from text is greatly improved with entity and anaphora resolution capabilities. However, entity and event anaphora resolution rely on (among other things) the semantic typing of the anaphor and its potential antecedents, particularly with sortal and event anaphora, as shown in (1) below.

(1) a. "For separation of nonpolar compounds, the pre-run can be performed with *hexane*$_i$; ... The selection of *this solvent*$_i$ might be considered .."

  b. [*p21*$_i$ inhibits the regulation of ...] ... [*This inhibitor*$_i$ binds to ...]

  c. [A *phosphorylates*$_i$ B.] ... [*The phosphorylation*$_i$ of B ...]

Strict UMLS typing presents a problem for our anaphora resolution algorithm (Castaño et al., 2002). For example, for the case of anaphora in (1a), the UMLS Metathesaurus types *hexane* as either 'Organic Chemical' or 'Hazardous or Poisonous Substance'. However, *solvent* is typed as 'Indicator, Reagent, or Diagnostic Aid'. In the UMLS Semantic Network, these semantic types are not related. Therefore the resolution of the sortal anaphora would fail, due to the type mismatch. The fact is that hexane is a solvent, and this is simply not reflected in UMLS.

Functional subtyping is also missing, as (1b) illustrates. This example shows a known protein (p21) being subsequently referred to as an 'inhibitor' (a functional class of proteins). This type does not exist in UMLS and the noun 'inhibitor' is merely typed as 'Chemical Viewed Functionally', while p21 itself is typed as 'Gene or Genome', AAPP, or 'Biologically Active Substance'. It is therefore difficult to discriminate p21 from other proteins (as potential antecedents) for the sortal anaphor "this inhibitor".

A related difficulty is encountered with event anaphora cases such as (1c), where an event nominal anaphor binds to a tensed event as its antecedent, both of which are of different types in the UMLS. Hence, the existing UMLS system does not allow for recognition of type-subtype relations of the kinds that are needed in order to identify anaphoric bindings in Medline texts.

Given these motivations, we have developed a set of techniques for "rerendering" an existing semantic ontology to satisfy the requirements of specific features of a (set of) application(s). For the present case (i.e., the UMLS and bio-entity and relation extraction), we identify candidate subtypes for inclusion in the type system by two means: (a) corpus analysis on compound nominal phrases that express unique functional behavior of the compound head; (b) identification of functionally defined subtypes derived from bio-relation parsing and extraction from the corpus. The results of rerendering are evaluated for correctness against the original type system, and against additional taxonomies, should they exist, such as the GO ontology. In our preliminary experiments, we had domain experts partially verify it aganst the Gene Ontology. Full automatic verification will be done in the future.

## 2. Semantic Rerendering

Many NLP tasks in the service of information extraction can benefit from more accurate semantic typing of the syntactic constituents in the text. As mentioned above, the semantic taxonomy available from UMLS is lacking in several respects. With specific applications such as content summarization, anaphora resolution, and accurate relation identification in mind, we describe how an existing type system can be systematically adapted to better serve these needs, using a technique we call *semantic rerendering*. Semantic rerendering is a process that takes as input an existing type system (such as UMLS) and a text corpus, and proposes refinements to the taxonomy on the basis of two strategies:

- *Linguistic Rerendering*: Syntactic and semantic analysis of NP structures in the text;

- *Database Rerendering*: Analysis of "ad hoc abstractions" from a database of relations automatically derived from the corpus.

In the first strategy, we use the syntax of noun groups to identify candidate subtypes to an existing UMLS type. For example, categories that are of interest to biologists but which are not explicitly represented in the type system are functional categories such as *phosphorylators*, *receptors*, and *inhibitors*. These are each significant categories in their own right but also have a rich number of subtypes as well, as illustrated in (2) below.

(2) 
```
CB(2) receptor
cannabinoid receptor
cell receptor
D1 dopamine receptor
epidermal growth factor receptor
functional GABAB receptor
gastrin receptororphan receptor
orphan nuclear receptor
major fibronectin receptor
mammalian skeletal muscle acetylcholine receptor
normal receptor
PTHrP receptor
protein-coupled receptor
ryanodine receptor
```

If individual proteins can be identified (i.e., semantically tagged) as belonging to a functionally defined class, such as *receptor*, then richer information extraction and textual binding is enabled.

There has been some recent research on extracting hyponym and other relations from corpora (Hearst, 1992; Pustejovsky et al., 1997; Campbell & Johnson, 1999; Mani, 2002). Our work extends the techniques described in (Pustejovsky et al., 1997) using more extensive corpus analytic techniques as developed in Pustejovsky & Hanks (2001).

### 2.1. Linguistic Rerendering

We first describe the linguistic rerendering procedure for inducing subtypes from corpus data, given an existing taxonomy such as the UMLS. We being by taking the strings classified as $<supertype>$ in the current type system. On the basis of their behavior in the corpus, we identify candidate subtypes, derived from an analysis of the structure of nominal compounds and clusters. We use the RHHR (*righthand head rule*, cf. Pustejovsky et al. (1997)) for compound nominals (CN) and create subtype $<head$-$of$-$CN>$ from the type of the head of CN. We then create a node for type $N'$ and insert it into the existing UMLS hierarchy.

More explicitly, the procedure for identifying candidate subtypes from the structure of nominal compounds is given below.

(1) Acquire corpus $C$.

(2) Apply existing type system $UMLS_1$ over $C$:
$TS\text{-}UMLS_1(C) = C_{S-Tag}$.

(3) Select from the resulting semantically tagged corpus $C_{S-Tag}$ all *NPs* with semantic tag $A$ with $\theta > \delta$, where $\theta$ is a measure of how interesting semantic type is for rerendering:

(4) For a given noun *N* that is the headword of a phrase with semantic tag $A$, propose *N* as *name* of a subtype of *S-Tag* $A$, $N' \sqsubseteq A$, if:

- N appears as head in a certain number of NPs of length $l \geq 2$;

- N falls under the threshold set for the headwords above, but is an LCS (longest common subsequence) of a number of syntactic heads that achieve it when combined [1];

- there is sufficient variation in words comprising the remainder of phrase (so as to exclude using collocations as subtypes).

(We will refer to the nodes inserted into the ontology at this stage *first-level extension*)

(5) Nouns in the residue of *NP* with *N* as head $\alpha$ as modifier can be proposed as subtypes of $\alpha N' \sqsubseteq N'$ (*second-level extension*).

---
[1] E.g. For AAPP, *oxidase* might not achieve the threshold by itself. However, it does when all headwords containing it as a subsequence are combined (i.e. *myeloperoxidase, peroxidase, deepoxidase*, etc.)

Further subcategorization of induced types, based on the analysis of modifiers within the nominal phrases, uses a combination of template filtering of noun phrases and the LCS (longest common subsequence) algorithm (Cormen et al., 1990). Notice that one must use different thresholds for headwords and modifiers (in step (4) or step (5) of the algorithm). However, at step (4), a candidate subtype may replicate exactly the parent node ($receptor \sqsubseteq Receptor$). In that case, first-level extension types must derived from subphrase analysis, but using the threshold established for step (4).

Once the candidate subcategories are selected, the next step is to obtain the instances for the induced subtypes. These instances and their type bindings can be identified from the corpus using a number of standard methodologies developed in the field for the expansion of ontology coverage (Hearst, 1992; Campbell & Johnson, 1999; Mani, 2002). For the moment, in the experiments we conducted, we used syntactic pattern templates to identify the strings that map to the proposed extensions of UMLS types (see examples in Table 1 below).

This procedure will result in differential depth of UMLS extension for functionally defined vs. naming categories. For example no strings should map to $\{head, neck, arm, leg\} \sqsubseteq <Body\ Location\ or\ Region>$, while string mappings are easily obtained for relational nouns such as $\{solvent, antibody, conjugate\} \sqsubseteq <Indicator, Reagent, or\ Diagnostic\ Aid>$.

### 2.2. Database Rerendering

The second strategy uses a database of biological relations constructed through the application of robust natural language techniques as outlined in Pustejovsky et al. (2002) and Castaño et al. (2002). Over this database, "ad hoc" categories are created by projecting statistically thresholded arguments. More formally, for a particular relation, a typed projection is obtained:

$$\pi X = \{X : T_1 | R(X, Y) \wedge T_1 \in UMLS_1\}$$

| $R$ | $X$ | $Y$ |
|---|---|---|
| phosphorylate | "TNIK" | "Gelsolin" |
| phosphorylate | "GSK-3" | "NF-ATc4" |
| phosphorylate | "IKK-beta" | "IkappaB" |
| ... | ... | ... |
| inhibit | "PD-ECGF" | "DNA synthesis" |
| inhibit | "BMP-7" | "terminal chondrocyte differentiation" |
| block | "DFMO" | "ODC activity" |
| abrogate | "Interleukin-4" | "hydrocortisone-induced apoptosis" |
| ... | ... | ... |

Table 2: A sample segment of relations database

The noun forms for such ad hoc categories are determined by checking each relation against the first-level extension subtypes derived through NP structure analysis as outlined above. Thus,

- For relation $R$ and each subtype $N' \sqsubseteq T_1$, associate $N'$ with $\pi X$ if $Sim(N, \pi R) > \epsilon$.

e.g. $Sim($ "kinase", "phosphorylate" $)$,
$Sim($ "inhibitor", "inhibit" $)$, etc.

Note that the ad hoc category created through projection of the relation's argument can be matched with the types obtained at the second-level of NP-based ontology extension.

The similarity measure is constructed as a weighted combination of string similarity (e.g. *LCS*-based score), and an integrated composite measure derived from the training corpus and the outside knowledge sources. The latter might use standard IR similarity measures on contexts of occurrence of $R$ and $N$ in Medline abstracts, in definitions of $R$ and $N$ in domain-specific MRDs (such as the On-line Medical Dictionary), etc. Thus, we have:

$$Sim(N, \pi R) = \\ = z_0 * \text{LCS-score}(N, \pi R) + \Sigma_{i=1}^{k} z_i * Sim_i(N, \pi R)$$

where $Sim_i(N, \pi R)$ is the similarity score derived from the source $i$, and $z_i$ is the weight assigned to the source $i$.

## 3. Methodology

### 3.1. Seed Ontology

The Unified Medical Language System (UMLS) which was used as the seed ontology has three components: the UMLS Metathesaurus, the UMLS Semantic Network, and the SPECIALIST Lexicon (UMLS Knowledge Sources, 2001). The UMLS Metathesaurus maps single lexical items and complex nominal phrases into unique concept IDs (CUIs) which are then mapped to the semantic types from the UMLS Semantic Network. The latter type taxonomy is what was used in the experimental applications of rerendering procedure. It consists of 134 semantic types hierarchically arranged via the 'isa' relation and interlinked by a set of secondary non-hierarchical relations. UMLS Metathesaurus in the UMLS 2001 distribution contains over 1.5 million string mappings.

In the Metathesaurus, multiple semantic type bindings are specified for many of the concepts. Due to this ambiguity of UMLS concepts and to a lesser extent, the ambiguity of the strings themselves, the mappings obtained from the Metathesaurus give a number of semantic types for each lexical item or phrase. We intentionally avoid superimposing any disambiguation mechanism on this typing information while applying it in corpus analysis. Since corpus-based derivation of subtypes uses a frequency cutoff, this ambiguity essentially resolves itself. For example, if a given lexical item is typed as both $T_1$ and $T_2$ in the seed ontology, and occurs as a headword in $> 1\%$ of nominal phrases typed as $T_1$, but in $< 1\%$ of nominal phrases typed as $T_2$, it will only be proposed as a candidate subtype of $T_1$. Thus, under the 1% cutoff, *isozyme*, which the seed UMLS types as either $Enzyme$ or $AAPP$, will only be identified as a good candidate subtype for $Enzyme$.

### 3.2. Corpus preprocessing with UMLS types

The experimental application of the rerendering procedure was conducted on a relatively small corpus of Med-line abstracts (around 40,000). Medline abstracts were tokenized, stemmed, and tagged. They were then shallow-parsed, with noun phrase coordination and limited prepositional attachment (only *of*-attachment) using finite-state techniques. The shallow parse was obtained using five separate automata each recognizing a distinct family of grammatical constructions, very much in the spirit of Hindle (1983), McDonald (1992) and Pustejovsky et al. (1997). The machinery used in preprocessing is described in more detail in Pustejovsky et al. (2002).

| Pattern Type | TEMPLATE | |
|---|---|---|
| apposition | "X, a Y inhibitor" | "X, the solvent |
| | "X, an inhibitor of Y" | "the solvent, X" |
| | "X, an inhibitor of Y" | "X, a common solvent for Y" |
| nominal compounds | "Y inhibitor" | "the solvent X" |
| definitional constructions | "X is an inhibitor of Y" | |
| aliasing constructions | "X (inhibitor of Y)" | "X (the solvent)" |
| | "an inhibitor of Y (X)" | "the solvent (X)" |
| enumeration | "Y inhibitors such as X, ..." | "solvents (e.g. X)" |
| | | "solvents, e.g. X" |
| | | "the following solvents: X, .." |
| relative clauses | "X which is an inhibitor of Y" | "the solvent used was X" |
| | | "X proved to be a suitable solvent" |
| adjuncts | | "in X and Y as solvents" |
| | | "X as solvent" |

Table 1: Sample syntactic patterns for string-to-semantic type mappings

Semantic type assignment of the resulting nominal chunks is determined through lookup as follows. Each noun phrase is put through a cascade of hierarchically arranged type-assignment heuristics. Higher priority heuristics take absolute precedence; that is, if a semantic typing is possible, it is assigned. In this implementation, we use the full UMLS semantic type hierarchy, including the mappings to both leaves and intermediate nodes.

During direct lookup, a string is assigned a given semantic type if the UMLS Metathesaurus contains a mapping of that string to a concept so typed. If a semantic type for the whole phrase is not found in UMLS, we attempt to identify its syntactic head using a modification of RHHR (*righthand head rule*), and determine the semantic type of the headword. For chunks with *OF*-attachment, i.e. phrases of the form, <NP-1> of <NP-2>, the lookup is first attempted on *NP-1* as a whole.

If the lookup on a particular prospective head fails, it is tested for a match with morphological heuristics recognizing semantically vacant categories, such as 'NUMERIC', 'ABBREVIATION', 'SINGLE CAPITAL LETTER', 'SINGLE LOWER-CASE LETTER', etc. These, and phrases headed by common words occurring in a non-specialized dictionary are filtered out. The last layer of heuristics applied to a prospective syntactic head successively attempts to strip a groups of suffixes and prefixes and perform lookup on the remaining stem.

### 3.3. Inducing candidate subtypes

In these initial series of tests, we experimented primarily with the first part of the rerendering procedure as it is outlined in Section 2.1. In the first stage of identifying the subtypes based on the syntactic analysis of noun phrase structure, a headword was considered a candidate subtype of type $T$ if it occured in more that 1% of all nominal chunks tagged as $T$. Note that the same chunk is frequently tagged with several UMLS types.

The candidate subtypes for the second (NP modifier-based) level of UMLS extension were identified using a combination of template and frequency-based filtering of noun phrases and the LCS (longest common subsequence) algorithm. Thus, for a given headword proposed as subtype at first level of extension (e.g., *kinase*) the LCS algorithm was run on all phrases with that headword that matched a certain template (e.g. <Indefinite Article> <Modifier>* N). The substrings that occurred in the corpus in more than a certain percentage of phrases with that headword were identified as candidate subtypes for insertion into the ontology at the next level. The cut-off threshold had to be kept very low for this series of experiments, as it was conducted over a relatively small corpus. In working with a larger corpus the thresholds are set separately for each template, so e.g. it is much higher for the unfiltered set of nominal compounds than for those occurring with an indefinite article. Frequency-based filtering involves discarding as potential candidates noun phrases with modifiers that occur frequently in separate non-specialized corpus, which allows to automatically discard phrases such as 'multiple receptors', 'specific kinase', etc. [2]

Identification of sample instances for the induced types was performed over shallow-parsed text using syntactic pattern templates. The definitional construction patterns were extracted using relation extraction machinery (see Pustejovsky et al. (2002) for details). It was applied to our test corpus and another sample set of Medline abstracts (approx. 60,000).

### 4. Results

Semantic typing over our sample set of Medline data produced type bindings for over 1 million noun phrases.

---
[2]Similar filtering was also applied to the first-level extensions

## 4.1. NP analysis-based subtypes

The choice of particular UMLS categories as supertypes for extension of the seed UMLS semantic type taxonomy is dictated by the particular natural language application. Semantic types given below are derived from nominal phrase analysis for some of the supertypes that have been used in anaphora resolution tasks (cf Castaño et al. (2002)). Each UMLS type is shown with the number of noun phrases of that type which occurred in our test corpus, followed by a list of derived candidate subtypes with their respective frequencies. The subtypes shown below were derived as described above in step 4 of the rerendering procedure specification in Section 2.1.

```
Enzyme 4724
     dehydrogenase 140
     protease 160
     reductase 73
     metalloproteinase 48
     isozyme 54
     oxidase 79
     phosphatase 111
     enzyme 1142
     kinase 741

Amino Acid, Peptide, or Protein 20830
     receptor 2444
     protein 4521
     peptide 947
     kinase 741
     cytokine 287
     isoform 412

Cell 16348
     macrophage 251
     clone 350
     neuron 1094
     lymphocyte 412
     fibroblast 257
     cell 11586

Cell Component 2508
     cytosol 84
     nucleus 469
     liposome 43
     organelle 40
     vacuole 35
     ribosome 28
     cytoskeleton 55
     dendrite 53
     cytoplasm 195
     soma 26
     granule 80
     chromatin 36
     microtubule 45
     chromosome 319
     axon 99
     microsome 132
```

Notice that the categories derived in this manner would include functionally defined types (e.g. *isoform*).

## 4.2. NP modifier-based extension (second-level)

As mentioned above, some of the UMLS extension candidates that are derived according to the procedure are replicas of the supertype category, e.g. $enzyme \sqsubseteq Enzyme$, or $receptor \sqsubseteq Receptor$. For example, among the lexical items tagged as *Receptor* in UMLS Metathesaurus, NPs headed by the word "receptor" comprise 87% of all NPs tagged as *Receptor* in our test corpus:

```
Receptor 2820
     integrin 91
     receptor 2444
```

The appropriate extensions to the comparable level within the type taxonomy in this case are derived from subphrase analysis. Thus, for the case of *enzyme*, the candidate subtypes so derived would be:

```
cytosolic enzyme
heterologous enzyme
male enzyme
metalloenzyme
multifunctional enzyme
proof-reading enzyme
proteolytic enzyme
rate-limiting enzyme
recombinant enzyme
rotary enzyme
tetrameric enzyme
```

These are identified at step 5 of rerendering procedure through a combination of template filtering of noun phrases and longest common substring identification. They are then added to the same level of the type taxonomy as all $N' \sqsubseteq Enzyme$ (see Figure 1).

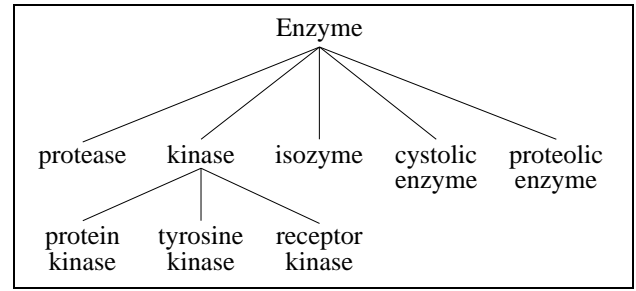

Figure 1: Extension subtree for $Enzyme$ (partial)

The results produced at this stage by the automated processing described above need further filtering before good subtype candidates can be identified. This can be achieved by fine-tuning the use of corpus frequencies, as well as type filtering of modifiers using the seed ontology type system. Table 3 below shows UMLS types for selected NP modifier-based subcategories of $receptor$.

## 4.3. Corpus-based identification of the instances of induced semantic categories

The rerendering procedure gives different results for different segments of the taxonomy, depending on the class of supertype category. Thus, for functionally defined semantic types, such as, "Chemical Viewed Functionally", or "Indicator, Reagent, or Diagnostic Aid", corpus-based derivation of instances for the induced subcategories is clearly much more feasible. Consider the first level extension types for the categories below:

```
Indicator, Reagent, or Diagnostic Aid 3424
     buffer 151
     conjugate 112
     stain 75
     agar 38
     antibody 1640
     indicator 373
     solvent 38
     tracer 53
     dye 95
     reagent 113
     nitroprusside 51
     hydrogen peroxide 58
```

| **Candidate Subtypes** $\alpha N' \sqsubseteq N'$ | **Seed UMLS Type for Modifier** $\alpha$ |
|---|---|
| cell surface receptor | 'Cell Component' |
| membrane receptor | 'Tissue' |
| adhesion receptor | 'Acquired Abnormality', 'Disease or Syndrome' |
|  | 'Natural Phenomenon or Process' |
| activation receptor | no type binding |
| contraction receptor | 'Functional Concept' |
| estrogen receptor | 'Steroid', 'Pharmacologic Substance', 'Hormone' |
| dopamine receptor | 'Organic Chemical', 'Pharmacologic Substance', |
|  | 'Neuroreactive Substance or Biogenic Amine' |
| adenosine receptors | 'Nucleic Acid, Nucleoside, or Nucleotide', |
|  | 'Pharmacologic Substance', 'Biologically Active Substance' |
| insulin receptor | 'Amino Acid, Peptide, or Protein', |
|  | 'Pharmacologic Substance', 'Hormone' |
| TSH receptor | 'Amino Acid, Peptide, or Protein', 'Hormone' |
|  | 'Neuroreactive Substance or Biogenic Amine' |
| EGF receptor | 'Amino Acid, Peptide, or Protein', 'Hormone', |
|  | 'Pharmacologic Substance', |
| transferrin receptor | 'Amino Acid, Peptide, or Protein', 'Biologically Active Substance', |
|  | 'Indicator, Reagent, or Diagnostic Aid', 'Laboratory Procedure' |
| **receptor** | **'Amino Acid, Peptide, or Protein', 'Receptor'** |

Table 3: UMLS Typing of modifiers $\alpha$ for some sample subtypes $\alpha N' \sqsubseteq N'$ for $N'$ =*receptor*

```
Chemical Viewed Functionally 3494
    inhibitor 1668
    prodrug 62
    basis 1075
    vehicle 107
    radical 144
    base 265
    pigment 36
    surfactant 36
Pathologic Function 17752
    impairment 383
    stenosis 274
    other 450
    illness 209
    problem 1133
    dysfunction 493
    block 244
    carrier 219
    inflammation 243
    pathogenesis 497
    cavity 273
    hemorrhage 180
    occlusion 266
    lesion 1820
    infarction 449
    regression 237
    pathology 242
    infection 1782
    complication 1248
    separation 320
    degeneration 180
    stress 487
```

Table 4 shows the derivation of instances for the categories induced through noun phrase analysis (step 5), using the definitional construction template. The first column shows the actual strings that get the new type binding as *kinase* (in blue) and their original UMLS types (in black). Notice that for many of the strings that can be so typed, the seed UMLS type is either generic $AAPP$ or the type binding is absent altogether.

If the candidate subtype is a valid semantic category, such corpus-based identification of instances should work equally well irrespective of the level at at which the induced type is inserted. For example, see Table 5 below for NP modifier extensions of *receptor*.

| cell-surface receptors: |
|---|
| **polycystin-1** is a *cell surface receptor* |
| **Fas** is a *cell surface* death *receptor* |
| **CD40** is a *cell surface receptor* |
| **CD44** is a *cell surface receptor* |
| **The scavenger receptor BI** is a *cell surface* lipoprotein *receptor* |
| membrane receptors: |
| **Neuropilin-1** is a trans*membrane receptor* |
| **APJ** is a seven trans*membrane* domain G-protein-coupled *receptor* |
| **HER2** is a *membrane receptor* |

Table 5: Sample semantic type instances derived with the definitional construction template for subtypes of *receptor*

## 5. Evaluation of Rerendering Procedure

The evaluation of the performance for rerendering essentially boils down to whatever improvement is produced in precision and recall for the client applications. However, in order to do an earnest evaluation of performance of the rerendering algorithm, we would need to run it on a much larger corpus. This would allow for better candidate choices for the portions of the procedure that have been plagued by sparsity (e.g., in NP modifier-based candidate subtype selection). But most importantly, it would increase the coverage in terms of instances for which the type bindings are produced in the new type system.

| | | | |
|---|---|---|---|
| RING3 | | is | a novel protein kinase |
| unknown | | | Amino Acid, Peptide, or Protein |
| Raf–1 | | is | a serine–threonine protein kinase |
| Amino Acid, Peptide, or Protein | | | Amino Acid, Peptide, or Protein |
| Bcr–Abl | | is | a tyrosine kinase |
| Gene or Genome | | | Amino Acid, Peptide, or Protein |
| Csk | | is | a cytoplasmic tyrosine kinase |
| unknown | | | Amino Acid, Peptide, or Protein |
| WPK4 | | is | a wheat protein kinase |
| unknown | | | Amino Acid, Peptide, or Protein |
| p59(fyn) | | is | a non–receptor tyrosine kinase of the Src family |
| unknown | | | Family Group |
| FER | | is | a volume–sensitive kinase |
| Intellectual Product | | | Amino Acid, Peptide, or Protein |
| The UL97 protein | | is | a protein kinase |
| Amino Acid, Peptide, or Protein | | | Amino Acid, Peptide, or Protein |
| Dbf2 | | is | a multifunctional protein kinase |
| unknown | | | Amino Acid, Peptide, or Protein |
| the JNK p54 isoform | | is | an ets–2 kinase |
| Amino Acid, Peptide, or Protein | | | Amino Acid, Peptide, or Protein |
| Tyk2 | | is | a Janus kinase |
| Amino Acid, Peptide, or Protein | | | Amino Acid, Peptide, or Protein |
| PYK2 | | is | an adhesion kinase |
| unknown | | | Amino Acid, Peptide, or Protein |
| The product of the HER2 / Neu oncogene | | is | a receptor tyrosine kinase |
| Gene or Genome | | | Amino Acid, Peptide, or Protein |
| ERK5 | | is | a novel type of mitogen–activated protein kinase |
| unknown | | | Amino Acid, Peptide, or Protein |
| H–Ryk | | is | an atypical receptor tyrosine kinase |
| unknown | | | Amino Acid, Peptide, or Protein |
| FixL | | is | a sensor histidine kinase |
| unknown | | | Amino Acid, Peptide, or Protein |

Table 4: Definitional construction template at work for the $N' = kinase$

### 5.1. Usability in natural language applications

One of the client applications for the experiments we report here is coreference resolution. The anaphora examples in (3) below illustrate the impact of using the derived types. Even the test corpus we used actually contained enough information to produce the type bindings for some of the strings used in (3).

(3)  a. "Assays were conducted in *chloroform, toluene, amyl acetate, isopropyl ether, and butanol*. ... In *each solvent*,"

   b. "The extracts were prepared separately in *methanol, ethanol, phosphate buffer saline (PBS), and distilled water* as part of our study to look at ... Our results have shown that *all four solvents* were ..."

   c. "A 47-year-old man was found dead in a factory where *dichloromethane (DCM)* tanks were stocked. He was making an inventory of t he annual stock of DCM contained in several tanks (5- to 8000-L capacity) by transferring *the solvent* into an additional tank with the help of compressed air."

   (emphasis added)

The seed ontology induces a type mismatch between the anaphor and the antecedent. For example, in (3c), the original type bindings are:

- *TS-UMLS*$_1$(solvent)= 'Indicator, Reagent, or Diagnostic Aid';

- *TS-UMLS*$_1$(dichloromethane)= { 'Organic Chemical', 'Pharmacologic Substance', 'Injury or Poisoning' }

The rerendered ontology allows the induced semantic type *solvent* $\sqsubseteq$ <*Indicator, Reagent, or Diagnostic Aid*> to be included in the type bindings for "dicloromethane".

### 5.2. Evaluation against existing ontologies

We performed some test evaluations of the second-level extension subtypes against the Gene Ontology. Despite the very modest side of our test corpus, we observed significant overlap in some categories. Thus, for example, the 388 second-level extension subtype candidates for *receptor*, 12% were identified as concept names in the Gene Ontology.

In general, the preliminary results of applying the first step of the rerendering procedure algorithm to the UMLS semantic type taxonomy appear quite encouraging. In the future, better automated methods for the evaluation of rerendering results against the existing ontologies must be developed. And most importantly, the utility and usefulness of the rerendering algorithm must be evaluated vis-a-vis achieving improvement in precision and recall for client NLP applications.